# Prediction of terephthalic acid (TPA) yield in aqueous hydrolysis of polyethylene terephthalate (PET)

Hossein Abedsoltan [a], Zeinab Zoghi [b], Amir H. Mohammadi [c]

[a] Chemical Engineering Department, The University of Toledo, Toledo, OH, USA.

habedso@rockets.utoledo.edu

[b] Electrical Engineering and Computer Science Department, The University of Toledo, Toledo, OH, USA.

zzoghi@rockets.utoledo.edu

[c] Discipline of Chemical Engineering, School of Engineering, University of KwaZulu-Natal, Howard College Campus, King George V Avenue, Durban 4041, South Africa

amir_h_mohammadi@yahoo.com   ,   amir.h.mohammadi2@gmail.com

## Abstract

Aqueous hydrolysis is used to chemically recycle polyethylene terephthalate (PET) due to production of high-quality terephthalic acid (TPA), the PET monomer. PET hydrolysis depends on various reaction conditions including PET size, catalyst concentration, reaction temperature, etc. So, modeling PET hydrolysis by considering the effective factors can provide useful information for material scientists to specify how to design and run these reactions. It will save time, energy, and materials by optimizing the hydrolysis conditions. Machine learning algorithms enable to design models to predict output results. For the first time, 381 experimental data were gathered to model aqueous hydrolysis of PET. Effective reaction conditions on PET hydrolysis were connected to TPA yield. The logistic regression was applied to rank the reaction conditions. Two algorithms were proposed, artificial neural network multi-layer perceptron (ANN-MLP) and adaptive network-based fuzzy inference system (ANFIS). The dataset was divided into training and testing sets to train and test the models, respectively. The models predicted TPA yield sufficiently where the ANFIS model outperformed. R-squared ($R^2$) and Root Mean Square Error (RMSE) loss functions were employed to measure the efficiency of the models and evaluate their performance.

1. Introduction

The growth of PET wastes has started simultaneously with the advent of PET synthesis and PET growing applications to manufacture plastic products. PET is mainly applied to produce packages, films, and automotive parts **[1-3]**. Water bottles and PET films are the typical wastes that are left in the environment after one-time usage **[4, 5]**. Although PET doesn't have any detrimental effect on human health, PET wastes have yet been considered obnoxious for the nature life **[6]**. Their negative impact on the marine life is reported **[7, 8]**. Also, landfilling, which is the conventional approach to eliminate the appearance of plastic wastes in the environment, has proved to be an insufficient approach due to the negative effects the plastic wastes have on the quality of the soil for the species living there **[9-12]**. Incineration is another approach to remove plastic wastes. However, it's not prominent since toxic gases will release, causing air pollution. Moreover, the weight of plastic wastes will reduce only 15% in average after incineration- meaning still we need to deal with considerable mass of remained ashes that are toxic, more dangerous than the plastic wastes **[13-16]**.

PET chemical recycling is the mere approach that can provide a sustainable process to reuse PET wastes. The main techniques to chemically recycle PET wastes are glycolysis, methanolysis, aminolysis, and hydrolysis **[17-21]**. PET hydrolysis produces terephthalic acid (TPA) and ethylene glycol (EG). The produced TPA has a high purity that can be furtherly purified and applied to re-produce PET **[22-24]**. There are three types of hydrolysis: neutral, acid, and alkaline. Neutral hydrolysis is conducted at high temperatures and pressures in water while acid and alkaline hydrolyses are conducted in considerably less severe conditions by an acid or a base as a catalyst **[24-26]**. Hydrolysis has been developing due to the versatile acids and bases that can be synthesized to depolymerize PET **[25, 27-30]**. The struggle is to better the reaction conditions to recycle PET. For instance, ionic liquids have been synthesized and applied as catalysts for PET recycling due to the low volatility and high recyclability **[31-34]**.

Artificial intelligence (AI) has recently opened new aspects in chemical and petroleum areas **[35, 36]**. There are reports on how to apply machine learning, the main branch of AI, to predict outcome(s) of a process. Recent focus is on applications of artificial neural networks (ANNs). The goal is to apply optimization algorithms to build a model that can predict outcome with an acceptably high precision and accuracy. This necessity is originated from the concept of understanding how a process may respond and what possible outcome(s) could be expected. It's also important to determine the effective input factors and measure their influence on the outcome(s) **[37-39]**. This will resolve the possible errors that may occur when the data is gathered by executing experimental procedures. It will also save time, energy, and material usage by presenting a clear path on how a process should be done to generate experimental data. Some examples are to predict the solubility in ionic liquids, degree of polymerization, and viscosity values **[40-45]**.

This study focuses on the effect of artificial intelligence upon the PET hydrolysis results in TPA yield. MLP-ANN and ANFIS models were designed to predict TPA yield, the output variable. The effective factors on PET hydrolysis were fed into the models as the input variables to make predictions. These variables effect TPA yield in experimental studies. For that, 381 samples were gathered from literature on aqueous hydrolysis of PET to prepare the dataset. $R^2$ and RMSE were determined as the two parameters to evaluate the efficiency of the models and compare their performance. To our knowledge, this is for the first time that the MLP-ANN and ANFIS models are applied to predict TPA yield in aqueous hydrolysis of PET-unscrewing a new chapter for the application of machine learning algorithms in chemical recycling of plastic wastes.

## 2. Materials and Methods

### 2.1 Dataset preparation

When the PET waste is mixed with other polymer wastes or non-polymer contaminants, it's needed to separate the PET waste before applying a hydrolysis technique. The process includes sorting, separating, washing, etc. **[24, 46-48]**. Afterwards, there's a one-type plastic waste, PET waste, with acceptable quantity of contaminants as the feedstock for the hydrolysis process. This strategy has been applied for mixed plastic wastes as to clarify the reaction path and to avoid the separation processes for the produced products **[49-51]**.

In addition, considering mixed waste PET or the solvent system type will cause to have multiple output variables in a dataset. This will create the multi-output learning problem for the dataset. Four challenges are raised for training a dataset with multiple output, velocity, veracity, volume, and variety **[52, 53]**. Regardless of how fast we could gather a dataset with multiple output, velocity, and how probable the gathered dataset could contain noise, missing values etc., veracity, a dataset with multiple output will face with volume and variety challenges anyways **[52-54]**.

Volume refers to the size of the output space. With the C number of output variables and R number of instances in each variable, the output spaces grow to CR. Thus, the computational cost and processing time increase in comparison to one dimensional output space, i.e., single output. Also, if the size of the output increases, formulating a winning strategy to split the dataset for training and testing the machine learning algorithms will become complicated **[53, 55]**. Since there is no general agreement on an effective split strategy, single-output learning methods were selected **[56]**. In addition to cost, time, and split strategy, a big size of output space increases the probability of data imbalance issue which is one of the main data mining issues **[57, 58]**.

On the other hand, variety indicates the format of output variables. The increase in number of output variables causes the structure of the output space to be more complicated. The complex output space requires to formulate the output dependencies, designate an appropriate loss function, and design a model compatible with multi-output problems **[52-54]**. A solution to the stated challenges is to break the dataset with multiple output into datasets with single outputs **[59, 60]**. In this work, the variables effecting the aqueous hydrolysis of PET were considered and the output variable was TPA yield **[30]**.

To prepare the dataset, TPA yield for aqueous hydrolysis of PET were gathered from literature. 381 data was gathered, and they were split into two sets, 90% training and 10% testing, by applying randomly stratified sampling algorithm. The input variables that are effective on PET hydrolysis were considered. The number of the input variables were ten which are reaction temperature (T), reaction pressure (P), PET size, PET amount, catalyst type, overall catalyst concentration, solution amount, reaction type, reaction time (t), and reaction heating/mixing condition. **Table 1** presents the input variables and the number of gathered data points along with the associated references.

**Table 1 The dataset along with the input variables, number of data points and references.**

| 1* | 2* | 3* | 4* | 5* | 6* | 7* | 8* | 9* | 10* | 11* | R* |
|---|---|---|---|---|---|---|---|---|---|---|---|
| 60 - 80 | 1.97 | A, 12.57 | 0.052 | f | 2 – 2.09 | 150 | $a_2$ | 0.5 – 1.5 | $a^3$ | 48 | **[25]** |
| 100 - 135 | 1 | A, 0 – 12.57 | 0.0153 | a | 7.5 | 25 | $a_1$ | 0.42 – 145 | $a^1$ | 29 | **[61]** |
| 70 - 100 | 1 | A, 0 – 0.071 | 0.008 | b | 7 - 13 | 50 | $a_1$ | 2 - 100 | $a^1$ | 41 | **[62]** |
| 150 - 190 | 1 | A, 0.018 – 0.035 | 0.001 | a | 3 – 9 | 25 | $a_1$ | 1 – 12 | $a^2$ | 58 | **[63]** |
| 145 | N/R | A, 0.79 – 78.54 | 0.016 | c | 0.03 | 30 | $a_1$ | 1 - 4 | $a^3$ | 18 | **[64]** |
| 170 - 190 | N/R | B, chips | 0.01 | d | 0.0031 | 20 | $a_1$ | 0.42 – 7.92 | $a^4$ | 35 | **[65]** |
| 40 - 90 | 1 | C, 8 x 8 | 0.104 | a | 119.6 – 168.8 | 4.6 – 4.9 | $a_1$ | 1 – 15 | $a^3$ | 60 | **[66]** |
| 120 - 160 | 1.7 - 4.6 | B, flakes | 0.078 | e | 0.87 – 3.47 | 90 | $a_2$ | 0.17 – 2.00 | $a^5$ | 24 | **[67]** |
| 220 | N/R | C, 0.33 x 0.33 | 0.0052 | g | 2.75 | 5 and 10 | $a_2$ | 0.03 – 2.00 | $a^4$ | 14 | **[68]** |
| 120 - 200 | N/R | B, flakes | 0.26 | g | 1.125 | 520.45 | $a_2$ | 1 – 7 | $a^3$ | 9 | **[69]** |
| 70 - 95 | 1 | B, flakes | 11.5 | f | 1.385 – 4.125 | 1500 | $a_2$ | 1 - 4 | $a^6$ | 37 | **[70]** |
| 300 - 385 | 296 | C, 0.33 x 0.33 | 0.011 and 0.008 | h | 0 | 15.3 and 20.22 | $a_3$ | 0.02 – 1.00 | $a^4$ | 9 | **[71]** |

* **1.** Reaction temperature (°C) range **2.** Reaction pressure (atm) range, N/R: Not Reported **3.** PET sample configuration: **A:** PET area range (mm$^2$), **B:** PET shape, **C:** PET size (L x W) (mm x mm) **4.** PET sample amount (mol) **5.** Catalyst Type: **a:** sulfuric acid ($H_2SO_4$), **b:** nitric acid ($HNO_3$), **c:** dual functional phase transfer catalyst [$(CH_3)_3 N (C_{16}H_{33})]_3[PW_{12}O_{40}$], **d:** Zinc Sulfate ($ZnSO_4$), **e:** potassium hydroxide [K(OH)], **f:** sodium hydroxide [Na (OH)] + tributylhexadecylphosphonium bromide, **g:** sodium hydroxide [Na (OH)], **h:** No catalyst **6.** Solution Concentration (M) range **7.** Solution Amount (mL) **8.** Reaction type: **a$_1$:** Acid, **a$_2$:** Alkaline, **a$_3$:** Neutral **9.** Reaction time (hr) range **10.** Reaction heating and mixing condition: **a$^1$:** Reflux, Magnetic Stirring, **a$^2$:** Autoclave, Shaking, **a$^3$:** Autoclave, Stirring, **a$^4$:** Microwave Irradiation, No Stirring, **a$^5$:** Mini-reactor, Stirring, **a$^6$:** Reflux, Mechanical Stirring **11.** Number of data points **R:** Reference number

## 2.2 Artificial Neural networks

The formation of artificial neural networks (ANNs) is inspired by the human brain structure. Their power in solving nonlinear problems has caused them being employed in pattern recognition, anomaly detection, optimization, decision making, classification, regression, etc. An ANN is comprised of a combination of single computational units, i.e., neurons that are interconnecting mutually **[72]**. A large number of weighted neurons are grouped into sets, called layers. These layers have been trained to approach problems by adjusting the interconnections as per input signals. Input signals have been transferred across the hidden neurons to an output signal, using either or both linear and nonlinear activation functions.

One of the most used feedforward ANN is multi-layer perceptron (MLP). A feed forward neural network is one of the kinds with one direction data flow mapping inputs into corresponding output(s). A simple MLP is comprised of the input layer, one hidden layer and the output layer. Each layer consists of neurons. The number of neurons in the input layer indicates the number of input variables, and the number of neurons in the output layer represents the number of output variables **[73]**.

The probability scores are compared to the ground truth by loss function in the last layer to identify the errors, and adjust the parameters, weights, and biases in every iteration **[74]**. The higher the error measured by loss function, the less reliable is the MLP model. To increase the reliability of the model a training algorithm, such as backpropagation, can apply a gradient descent to loss function to find the optimal values for weights and biases, and to transfer it back to the earlier layers. Forward and backward passes repeat until there is no change seen in the performance of MLP by updating the parameters **[75]**. **Fig. 1** exhibits the schematic diagram of a two-layer MLP that was used in this work.

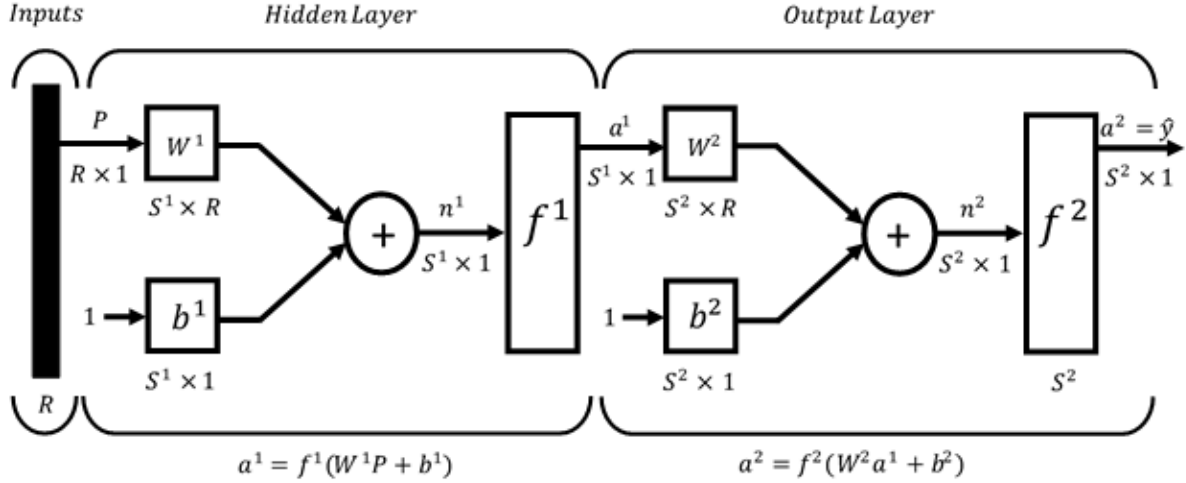

**Fig. 1 Schematic diagram of two-layer MLP**

$P$ represents a single vector of inputs with $R$ elements. The input signals are passed on to the next layer, going to the weight matrix $W$ with $R$ columns and $S$ rows. $S$ indicates the number of neurons in each layer. The constant 1 multiplies to the two scaler biases, $b^1$ and $b^2$. The net input $n$, which is $WP + b$, enters to the activation function. The activation function is applied on the net input and it generates the output $a$, which is $f(W^1P + b^1)$ in the hidden layer and $f(W^2 f(W^1P + b^1) + b^2)$ in the output layer. The commonly used activation functions are linear and Tanh, which were applied for the MLP model.

Linear activation function takes the inputs $a$ and multiplies by a constant gradient $c$. The descent continues down to the constant gradient. This activation function **(1)** is mostly utilized in the output layer.

$$f(x) = ca \tag{1}$$

tanh is the relationship between the hyperbolic sine and cosine **(2)** that is widely used for the input layer.

$$tanh(n) = \frac{e^n - e^{-n}}{e^n + e^{-n}} \tag{2}$$

The error associated with not trained or partially trained MLP model is calculated by means of loss function in the output layer. The most popular loss function for regression problem is Means-Squared Error, determined by **(3)**.

$$E = \frac{1}{n}\sum_{i=1}^{n}(y - \hat{y})^2 \tag{3}$$

$n$ represents the number of samples that the MLP model has been trained with. $y$ is the ground truth and $\hat{y}$ is the predicted value. The gradient of the loss function with respect to the weight $W$ is identified as shown in **(4)**.

$$\delta^o = (y - \hat{y}) \tag{4}$$

Where $\delta^o$ represents the error in the output layer. This error is back propagated to the hidden layer to obtain the error associated with that layer by **(5)**:

$$\delta^h = (1 - tanh(n)^2)\sum_{i=1}^{N} W_i^2\, \delta^o \tag{5}$$

Where $1 - tanh(n)^2$ is the derivative of the activation function in hidden layer. $W^2$ refers to the weight of the nodes in the hidden layer that are connected to the node in the output layer. $N$ represents the number of nodes in the hidden layer. The weights of the nodes in the output layer and the hidden layer are updated by **(6) and (7)**, respectively.

$$W^2 = W^2 - \eta\delta^o tanh(n) \tag{6}$$

$$W^1 = W^1 - \eta \delta^h P \tag{7}$$

Where $\eta$ indicates the learning rate, deciding how fast the MLP learns.

*2.3 ANFIS*

Adaptive network-based fuzzy inference system (ANFIS), which was first proposed in 1993, is a hybrid method that inherits both human-like rationale of fuzzy logic system (FLS) and learning power of ANNs **[76, 77]**. Fuzzy-based models are capable to formulate mapping rules that represent a relationship between the input and the output values. These rules are comprised of linguistic terms as it is utilized to decide in uncertain environment. While neural networks can learn from the data and tune the membership function using its backpropagation optimization algorithm, lacking a standard approach in the fuzzy-based models to map human knowledge into a set of rules and absence of a powerful method to tune the membership functions, make them unable to digest the facts about the data **[78]**. So, the ANFIS model with the power of both ANN and fuzzy inference system were used in this study.

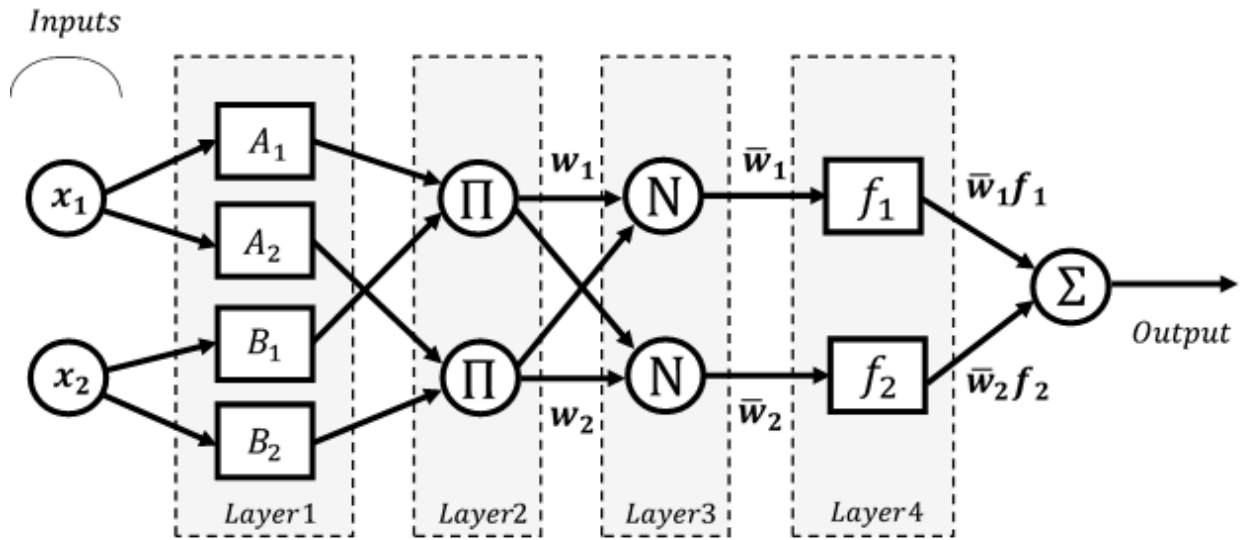

**Fig. 2 Schematic diagram of ANFIS**

The structure of the adaptive network that relies on the fuzzy inference system is built up of five layers. It is assumed that the network has two input variables $x_1$ and $x_2$, and one output variable y. **Fig. 2** shows the basic structure of adaptive network. It is constructed from fuzzification, input membership function, fuzzy rules, defuzzification, and output layers **[79]**.

The fuzzification layer (Layer 1) estimates how much input $x_i$ satisfies quantifiers $A_i$ and $B_i$ which are the linguistic labels of fuzzy set $i = \{1,2,...,n\}$. $i$ represents the $ith$ node that the variable $x$ is fed into and $n$ is the number of nodes in the first layer. This estimation was done by Gaussian membership function, presented in **(8)**.

$$\mu A_i(x_1) = \mu B_i(x_2) = f(x; a, b, c) = \frac{1}{1 + (\frac{x-c}{a})^{2b}}, i = \{1,2,...n\} \quad (8)$$

Where $a$ determines the width of the curve, $c$ represents the mean of the peak, and $b$ indicates the height or slope of the curve. A membership function decides on the membership degree of set of input variables. An input can be partially or fully a member of the fuzzy set or they may have no membership.

In Layer 2, each node in membership layer yields the product (logical conjunction) of incoming signals that remains unchanged from Layer 1. The outgoing signals ($w_i$) indicate the firing strength of a rule as shown in **(9)**.

$$w_i = \mu A_i(x_1) \times \mu B_i(x_2), i = \{1,2,...n\} \quad (9)$$

In Layer 3, the ratio of every incoming signal represents the rule's firing strength to the sum of the entire rules' firing strengths, which is computed in fuzzy rules layer.

$$\overline{w}_i = \frac{w_i}{w_1 + w_2}, i = \{1,2, \ldots n\} \tag{10}$$

For Layer 4, also known as defuzzification layer, the weighted value of rules is estimated. Individual adaptive nodes in this layer are fully connected to the corresponding normalization node, coming from fuzzy rule layer.

$$Output\ of\ defuzzification = \overline{w}_i f_i, i = \{1,2, \ldots n\} \tag{11}$$

Where $f_i$ represents the function that is implemented if the following fuzzy If-Then rules are true where the rules obey Takagi and Sugeno fuzzy inference systems are defined as follow **[80]**:

Rule 1: $If\ x_1\ is\ A_1\ and\ x_2\ is\ B_1, then\ f_1 = p_{10} + p_{11}x_1 + p_{12}x_2$

Rule 2: $If\ x_1\ is\ A_2\ and\ x_2\ is\ B_2, then\ f_2 = p_{20} + p_{21}x_1 + p_{22}x_2$

Where $A_1$, $B_1$, $A_2$, and $B_2$ are the linguistic labels of fuzzy sets and $p_{10}, p_{11}, p_{12}, p_{20}, p_{21}\ and\ p_{22}$ represent the consequent parameters.

The last layer (Layer 5), known as the output layer, is comprised of one node that computes the sum of the entire outputs, coming from the nodes in the defuzzification layer.

$$Output\ of\ ANFIS = \sum_i \overline{w}_i f_i \tag{12}$$

The output is eventually evaluated by applying backpropagation to tune the parameters of membership function. This evaluation has been repeated until adaptive network reflects no changes in performance.

*2-4 Data preprocessing*

The dataset was comprised of 381 samples and 11 variables containing 10 input variables and 1 output variable. Four number of the input variables are categorical and the remaining 6 variables are numerical. The categorical input variables were transformed to numerical which is the acceptable format for the normalization, MLP-ANN and ANFIS algorithms. We employed data normalization on the numerical input features to bring them all to the same range by using Power Transformer. Following the data normalization, a sampling technique was employed to split the dataset into training and validation-testing subsets. The sampling technique was composed of data shuffling and k-fold cross-validator followed 80:20 rule **[81]**.

The value of k for k-fold cross-validation was set to 5. The reason behind this value was that k-fold cross-validator went by 80:20 rule. According to this rule, the desirable size for training and validation-testing subsets should be 305 and 76, respectively. Using k-fold cross-validation, the size of the validation-testing subset was determined by dividing the number of entire samples by the number of the folds. Thus, the value of 5 for k, split the data into the desirable size (381/5=76). Since the size of validation-testing subset was set to 76, the remaining 305 samples were kept for training subset.

In this resampling method, the dataset was shuffled in each fold and 80 percent of the data samples were randomly selected and formed the training subset. The validation-testing subset was prepared by 20 percent of the data samples. Then, it was split into two separate validation and testing subsets, half formed the validation set and the other half formed the testing set. To find the best split, containing a good amount of information gathered from each experimental data, the split was repeated up to 4 times in each fold. This formed 20 training and validation-testing subsets at the end of 5 folds. It means that the data was reshuffled 4 times in each fold and made 4 different training and validation-testing subsets.

In each iteration, the validation-testing subset was reshuffled and randomly split into two subsamples of validation and testing with the size of 39 and 37, respectively. Once the validation and testing subsets were formed, the MLP algorithm was trained prior to ANFIS to find the best train-test split using validation subset and to decide upon the number of neurons in hidden layer. Therefore, the testing subset kept unseen to assess the performance of final MLP-ANN and ANFIS models on the test subset.

To find the best split, the reliability of the model was validated when the trained MLP-ANN model was implemented on the corresponding validation subset while the size of the hidden layer varied from 2 to 25 in each fold. The size of the hidden layer deviated from 2 to 25 to choose the optimum number of neurons in hidden layer. This process is discussed in detail in the next section. The correlation coefficient ($R^2$) was calculated for every subset of training and validation. **Fig. 3** demonstrates $R^2$ when MLP with 21 hidden neurons was implemented on training and validation subsets. These values are obtained through 5 folds of cross-validation. Since resampling method was repeated 4 times in each fold, MLP-ANN was implemented 4 times and accordingly $R^2$ was calculated for 20 different subsets.

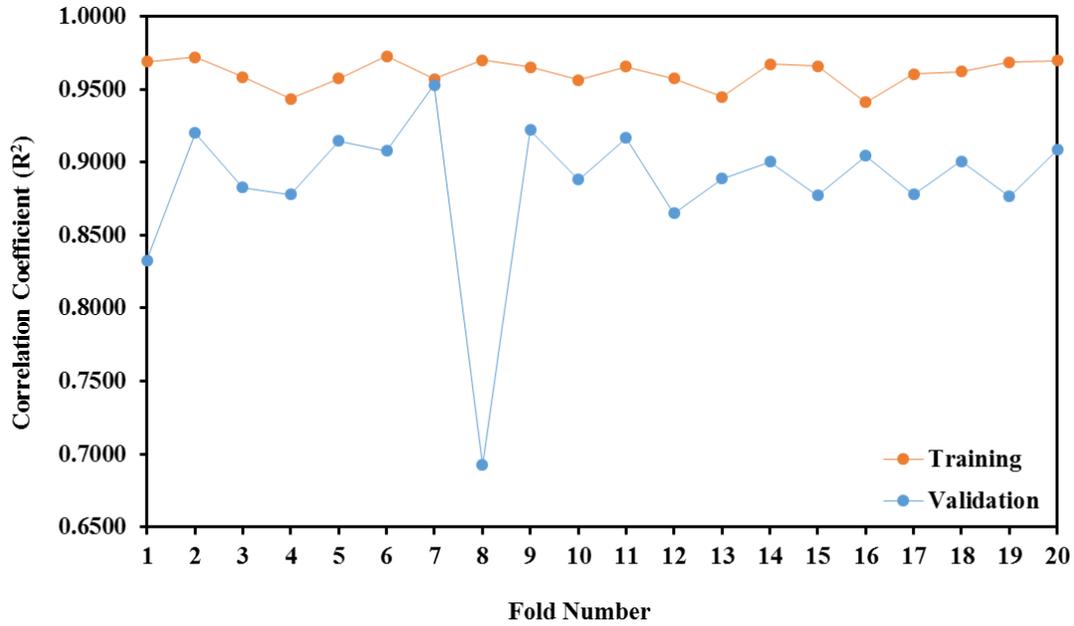

**Fig. 3 Correlation coefficient values ($R^2$) for training and validation phases**

As confirmed by the graph, the seventh iteration represents the best train-test split. In this iteration, the trained MLP-ANN model was applied on the validation subset with 21 neurons in the hidden layer. The model delivered the correlation coefficient with the values of 0.9572 and 0.9529, when it was implemented on the training and validation subsets, respectively. The $R^2$ value is acceptable in the training phase and it reaches its highest level in the validation phase, comparing to the other folds. Also, the small variance of the $R^2$ value in the training and validation phases indicates no overfitting and underfitting issues that makes the seventh test and train splits highly reliable to be fed into the machine learning models.

## 3. Results and discussion

The dataset was initially analyzed to determine the informative variables for the machine learning models. Each input variable was individually compared to the output variable to determine whether there is a strong relationship. The level of strength was measured using p-value. P-value is a metric in machine learning area to find if a combination of input variables is statistically important to effectively train a machine learning model **[82-85]**. P-value refers to the hypothesis of the significance level. The null hypothesis is that the input variable is not significant. So, the higher the p-value is, the more the hypothesis is correct.

Two univariate feature selection algorithms were applied to calculate the p-values. Continuous input variables are reaction temperature, reaction pressure, PET sample amount, overall catalyst concentration, solution amount, and reaction time. Pearson's algorithm was implemented in this case as both the input and the output variables are continuous **[86, 87]**. On the other hand, PET sample configuration, catalyst type, reaction type, and reaction heating and mixing condition are categorical. Kruskal-Wallis test was implemented to calculate the p-values for the categorical input variables **[88]**. **Table 2** presents the p-value scores calculated for the input variables of the dataset. All the calculated p-values were less than 0.05 which confirms the considered input variables were significantly important **[89-91]**. Thus, the entire input variables were considered for feature ranking and modeling.

**Table 2 Calculated p-values for the input features of the dataset**

| Input Feature | P-value |
|---|---|
| PET sample configuration | < 0.001 |
| Catalyst type | < 0.001 |
| PET sample amount (mol) | < 0.001 |
| Reaction heating and mixing condition | < 0.001 |
| Solution amount (mL) | < 0.001 |
| Reaction temperature ($^oC$) | < 0.001 |
| Reaction pressure (atm) | < 0.001 |
| Reaction type | 0.008 |
| Reaction time (h) | 0.013 |
| Overall catalyst concentration (M) | 0.047 |

To calculate the level of importance of each input variable, the logistic regression was implemented on the dataset **[92, 93]**. The logistic regression assigns an individual weight, also known as regression coefficient, to each input variable to measure its importance on the output prediction. Then, the regression coefficients were used to compute the rank of each input variable **[94-97]**. **Fig. 4** presents the ranking. The reaction temperature and the catalyst type are the most and the least important input variables on effecting TPA yield in aqueous hydrolysis of PET.

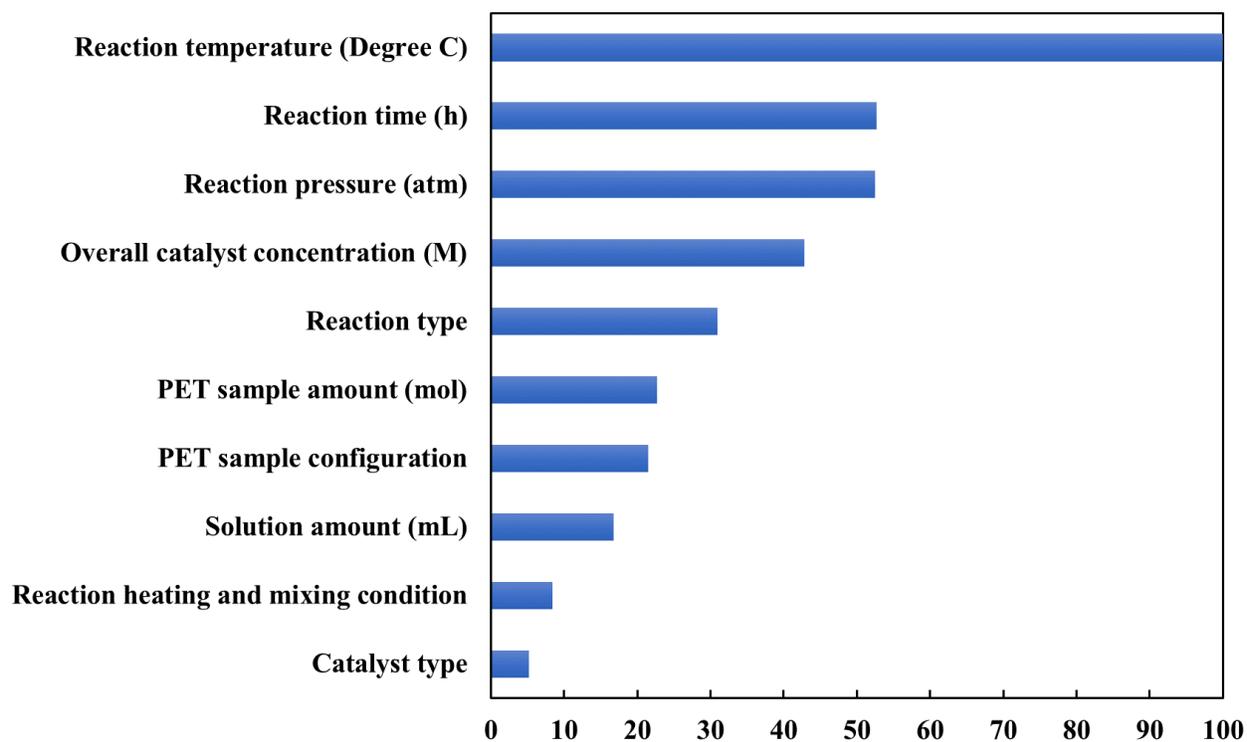

**Fig. 4 Rank of the effect of input variables on TPA yield, determined by logistic regression**

The 10 input variables were fed into the machine learning models for TPA yield prediction. 381 data samples were gathered from literature. 305 experimental data samples were held to form the training subset, 39 numbers of the remaining 76 data samples were taken for the validation subset, and the residual 37 data points were kept for the testing subset. The validation subset was used to identify the optimum train-test split ratio in the preprocessing phase. It was used simultaneously to determine the optimum number of neurons in the hidden layer. On the other hand, the testing subset was kept unseen and unemployed during the process of training and validation. This subset was used to estimate the predictive power of the final MLP-ANN and ANFIS models.

The number of neurons in hidden layers of MLP-ANN is one of the major hyper-parameters that builds the architecture of the network. It plays an important role in increasing the effectiveness of the models and decreasing the loss on the training subset. Too few hidden nodes reduce the performance of the model and result in underfitting issue while too many hidden nodes lead to the overfitting issue and poor model performance. A trial-and-error approach was utilized to decide on the optimum number of hidden nodes **[98]**. In this approach, a range of number of neurons were selected and the MLP-ANN model was implemented when the number of neurons varied in the range. The range was set from 2 to 25. The optimum number of neurons was gained where the predictive power of MLP-ANN model was improved and the $R^2$ values tend to be very close to each other in the training and validation phases.

**Fig. 5** shows $R^2$ values calculated from the predicted outcome by MLP-ANN when it was implemented on the optimum training and validation subsets. The orange square points represent the $R^2$ values for the training subset and the green square points represent the $R^2$ values for the validation subset. In **Fig. 5**, the $21^{st}$ orange and green points indicate the acceptable $R^2$ values. Also, the points are located on the nearest position across each other which means there are no overfitting and underfitting issue when the MLP-ANN model was applied with 21 hidden neurons on the validation subset. Therefore, the number of hidden neurons was set to 21 which is the optimum number of nodes for the hidden layer.

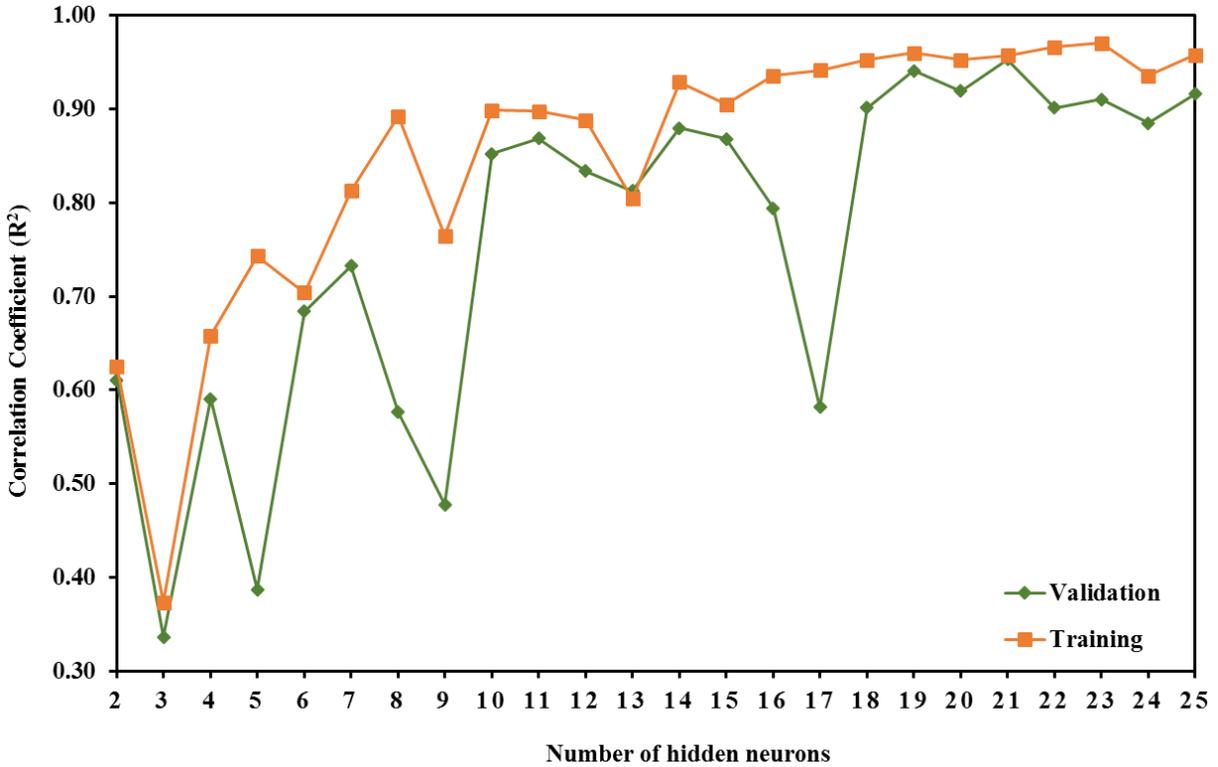

**Fig. 5 $R^2$ values for the validation and the training subsets, determined by MLP-ANN**

The performance of the MLP-ANN model with 21 neurons in hidden layer was evaluated by cross-checking $R^2$ and root mean square error (RMSE) values when the model was run on the validation subset. **Fig. 6** and **Fig. 7** demonstrate the results for $R^2$ and RMSE, respectively. In these figures, y-axis represents the $R^2$ and RMSE and x-axis indicates the range of the values that were selected as the number of neurons for MLP-ANN. **Fig. 6** shows that the highest $R^2$, which is 0.9529, delivers from an MLP-ANN model with 21 neurons in the hidden layer. In support of the results shown in **Fig. 6**, **Fig. 7** illustrates that the MLP designed with 21 neurons in the hidden layer generated the RMSE with the value of 6.95 which is a small error in comparison with the rest of the number of neurons.

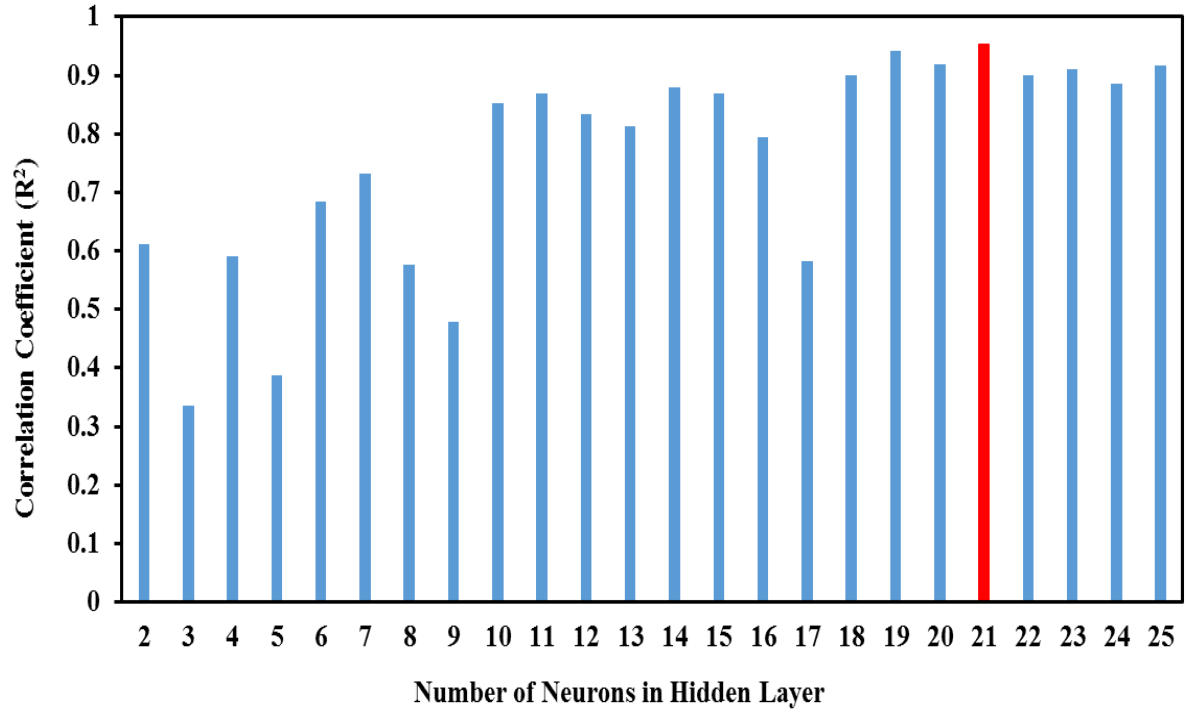

**Fig. 6 $R^2$ variation with respect to the number of neurons for the validation subset**

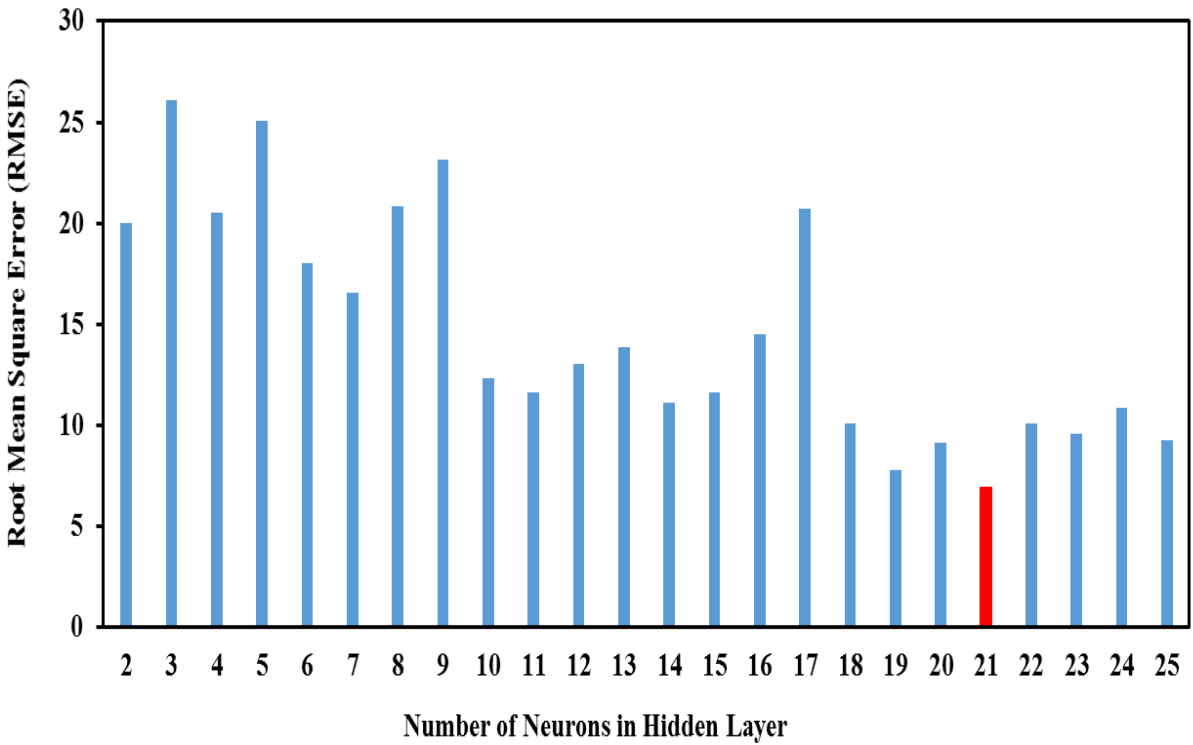

**Fig. 7 RMSE variation with respect to the number of neurons for the validation subset**

Once the size of hidden layer for MLP-ANN was determined, the hyper-parameters for MLP-ANN and ANFIS were tuned to optimize the architecture of the final models. **Table 3** represents optimal hyper-parameters considered for the MLP and ANFIS models.

**Table 3 Details of trained ANN with MLP and ANFIS for the prediction of TPA yields**

| Hyper-Parameter | Model Name | Value |
| --- | --- | --- |
| Hidden Layer Activation function | MLP-ANN | tanh |
| Output Layer Activation function | MLP-ANN | Linear |
| Optimization method | MLP-ANN | BP + L-BFGS |
| Hidden Layer Size | MLP-ANN | 21 |
| Input Layer Size | MLP-ANN | 10 |
| Output Layer Size | MLP-ANN | 1 |
| Number of Maximum Iteration | MLP-ANN, ANFIS | 200 |
| Optimization Method | ANFIS | BP + Least Square |
| Decrease rate | ANFIS | 0.9 |
| Increase rate | ANFIS | 1.2 |
| Initial Step | ANFIS | 0.0001 |
| Number of Epoch | ANFIS | 2 |
| Range of Influence | ANFIS | 0.9 |
| Squash Factor | ANFIS | 1.2 |
| Acceptance Ratio | ANFIS | 0.5 |
| Rejection Ratio | ANFIS | 0.2 |

After the hyper-parameters were tuned, MLP-ANN and ANFIS were trained with the training subset. Both models were trained and tested on the same training and testing subsets to accurately compare the results achieved by the MLP-ANN and ANFIS models, **Fig. 8 and Fig. 9** show the comparison of the predicted outcome generated by the MLP-ANN and ANFIS models against the final output at the training and testing phases. **Fig. 8 (a)** shows both the data points from real TPA yield and the data points from the predicted TPA yield, gained from MLP-ANN, when the trained model was implemented on the training subset. **Fig. 8 (b)** indicates the same data points when the MLP-ANN model was applied on the testing subset. The blue points present the real output, and the red points depict the predicted output. Similarly, in **Fig. 9 (a)**, the blue triangle points present the real TPA yield, and the red ones show the predicted TPA yield while the trained ANFIS model was implemented on the training subset. **Fig. 9 (b)** indicates the same data points when the ANFIS model was run on the testing subset.

In these figures, the difference between the red points and the blue points shows how well the models trained and how well they predicted the output. Comparing the outputs generated by MLP-ANN and ANFIS, the red points are closer to the blue points in **Fig. 9 (b)** for both the training and testing subsets. The small difference between the red points and the blue points in this figure indicates that the ANFIS model performed better than the MLP-ANN model.

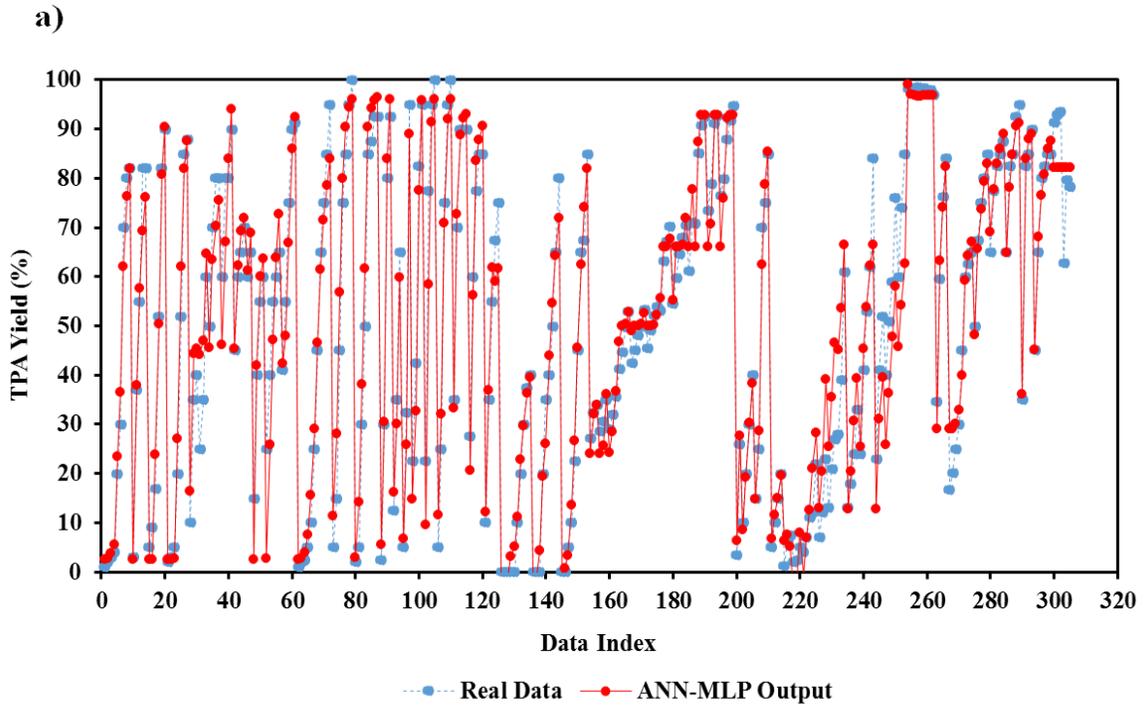

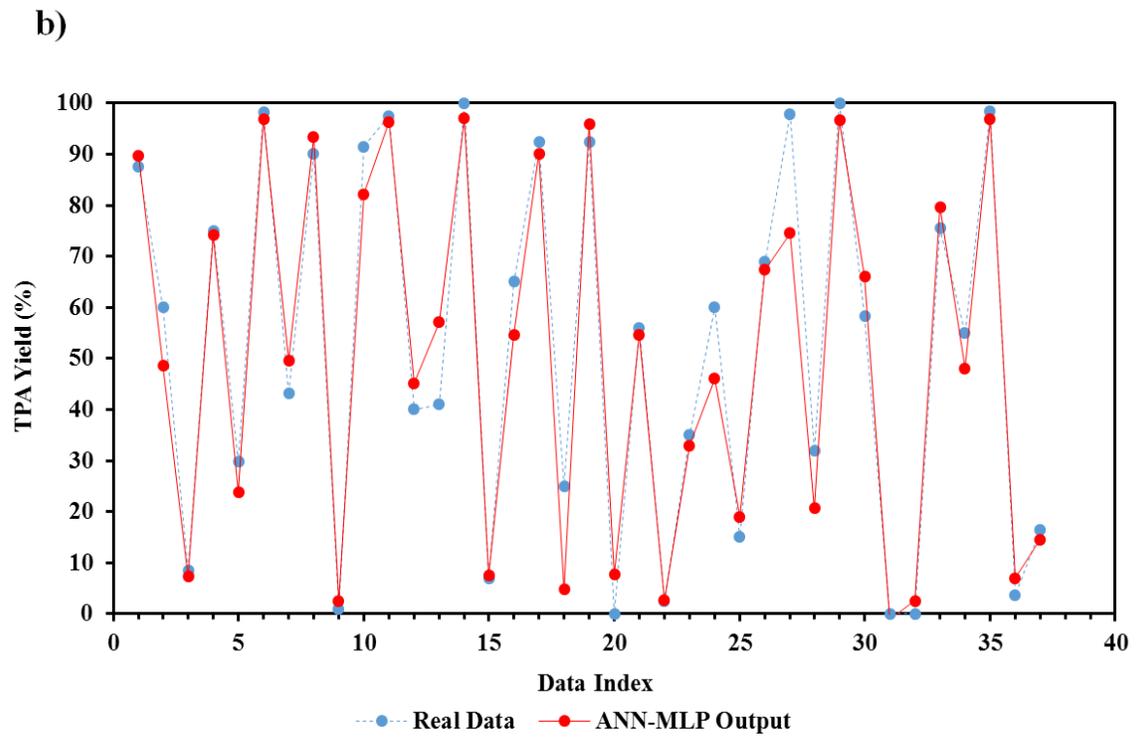

**Fig. 8 Real versus ANN-MLP predicted TPA yield (%): a) training set, b) testing set**

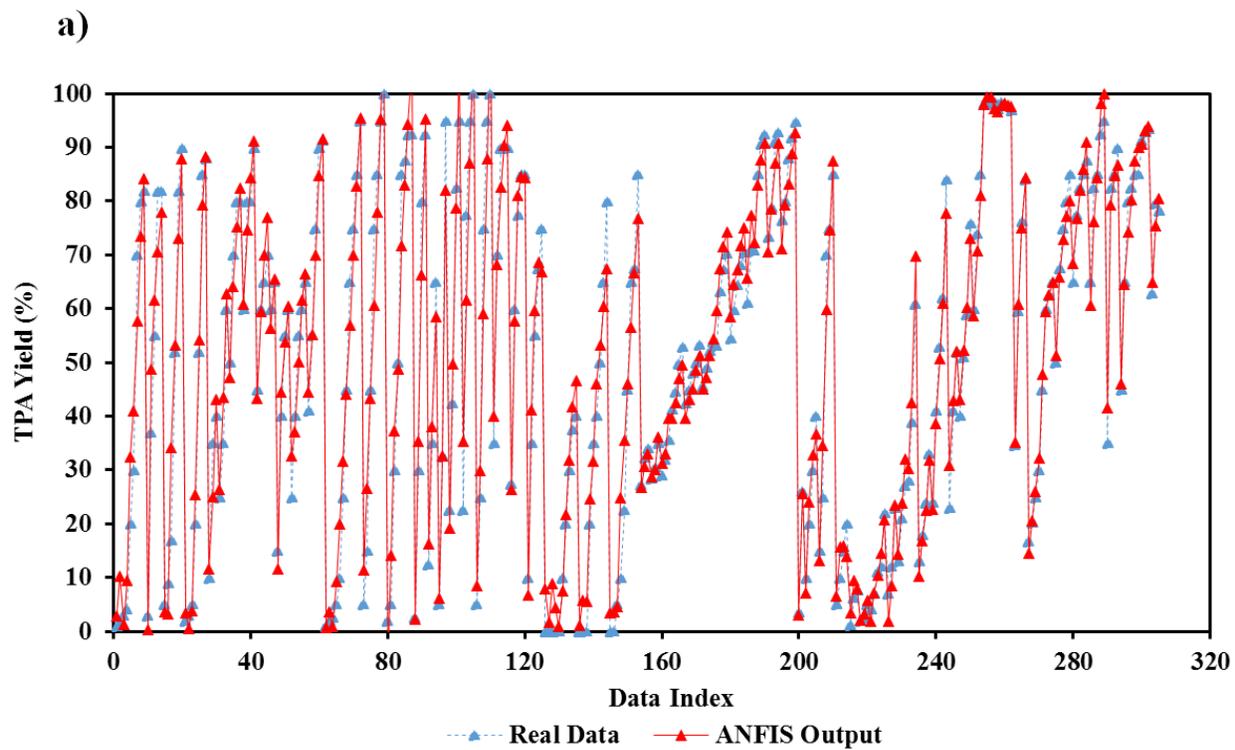
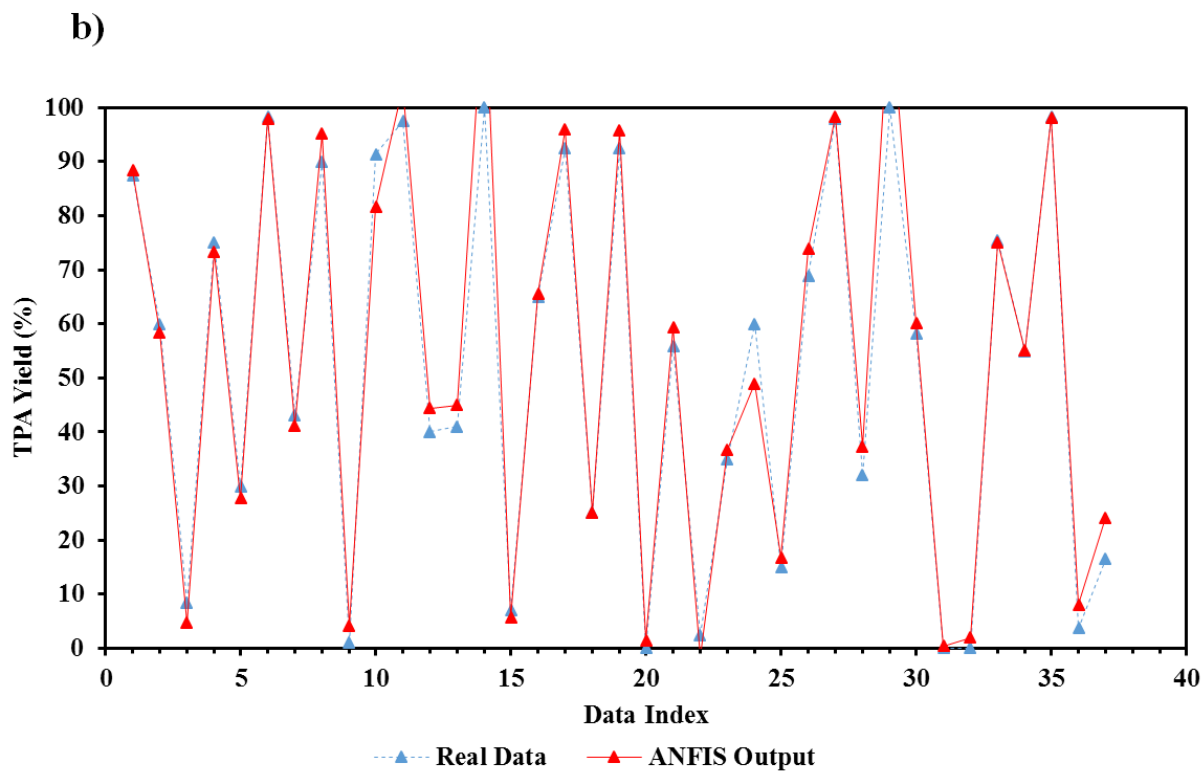

**Fig. 9 Real versus ANFIS predicted TPA yield (%): a) training set, b) testing set**

$R^2$ and RMSE calculations were conducted to do a comprehensive evaluation. $R^2$ and RMSE are the most known metrics used to accurately evaluate the precision of the predictions in regression models. Thus, they were applied in this study to measure the predictive power of the models as follow:

$$R^2 = 1 - \frac{\sum_{i=1}^{N}(y_i - \hat{\mu}_i)^2}{\sum_{i=1}^{N}(y_i - \bar{y}_i)^2} \tag{13}$$

$$MSE = \sqrt{\frac{1}{N}\sum_{i=1}^{N}(y_i - \hat{\mu}_i)^2} \tag{14}$$

Where N indicates the entire number of data points, $y_i$ is the ith actual data point, $\hat{\mu}_i$ is the ith predicted output, $\bar{y}_i$ is the average of the actual data points.

The $R^2$ value calculated by the MLP-ANN and the ANFIS implementations on the training and testing subsets are shown in **Fig. 10 and Fig. 11**, respectively. The closer the data points come to form a straight dashed line, the more probable the model accurately predicts the output. **Fig. 10 (a)** shows the $R^2$ values when the trained MLP-ANN model was run on the training subset. The same result is shown in **Fig. 10 (b)** while the trained MLP-ANN model was applied on the testing subset. Similarly, in **Fig. 11 (a) and Fig. 11 (b)**, the predictive power of ANFIS is evaluated using $R^2$ when the trained model was implemented on the training and testing subsets, respectively. In these figures, the values of the real TPA yields are shown against the values of predicted TPA yields. In **Fig. 11**, the points are scattered around the identity line both for the training and testing subsets, while the points lie farther against the identity line (straight dashed line) in **Fig. 10**. It indicates that the ANFIS model was more powerful in TPA yield prediction in comparison with the MLP-ANN model. The $R^2$ values of 0.9541 and 0.9738 were measured using the generated outcome achieved from the MLP-ANN and ANFIS models, respectively, when they were implemented on the testing subset. Also, the MLP-ANN model makes 7.81 and the ANFIS model makes 6.54 error. The ANFIS model delivers the satisfactory $R^2$ and less error while the lower $R^2$ value and the higher RMSE for MLP-ANN indicates that the predicted accuracy of MLP-ANN is lower than ANFIS. Thus, the ANFIS model outperformed the MLP-ANN model.

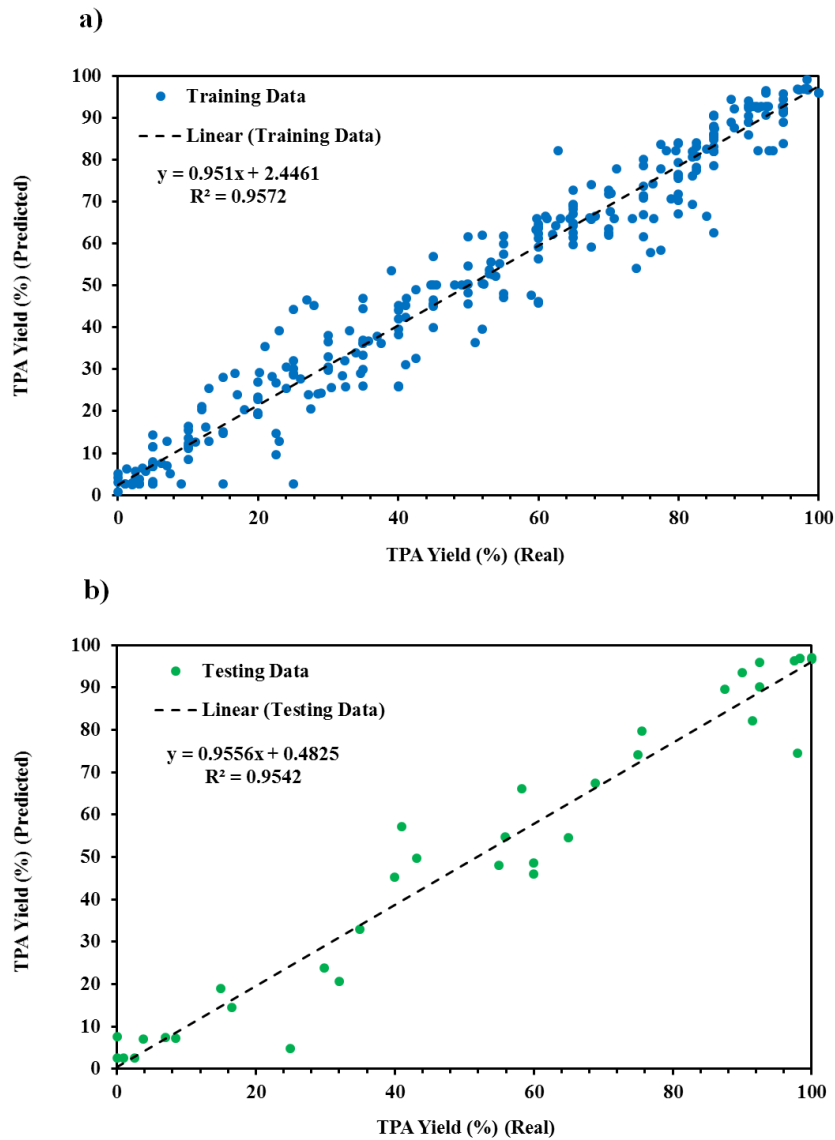

**Fig. 10 Regression plots for the TPA yield (%) prediction applying MLP-ANN model: a) training set, b) testing set**

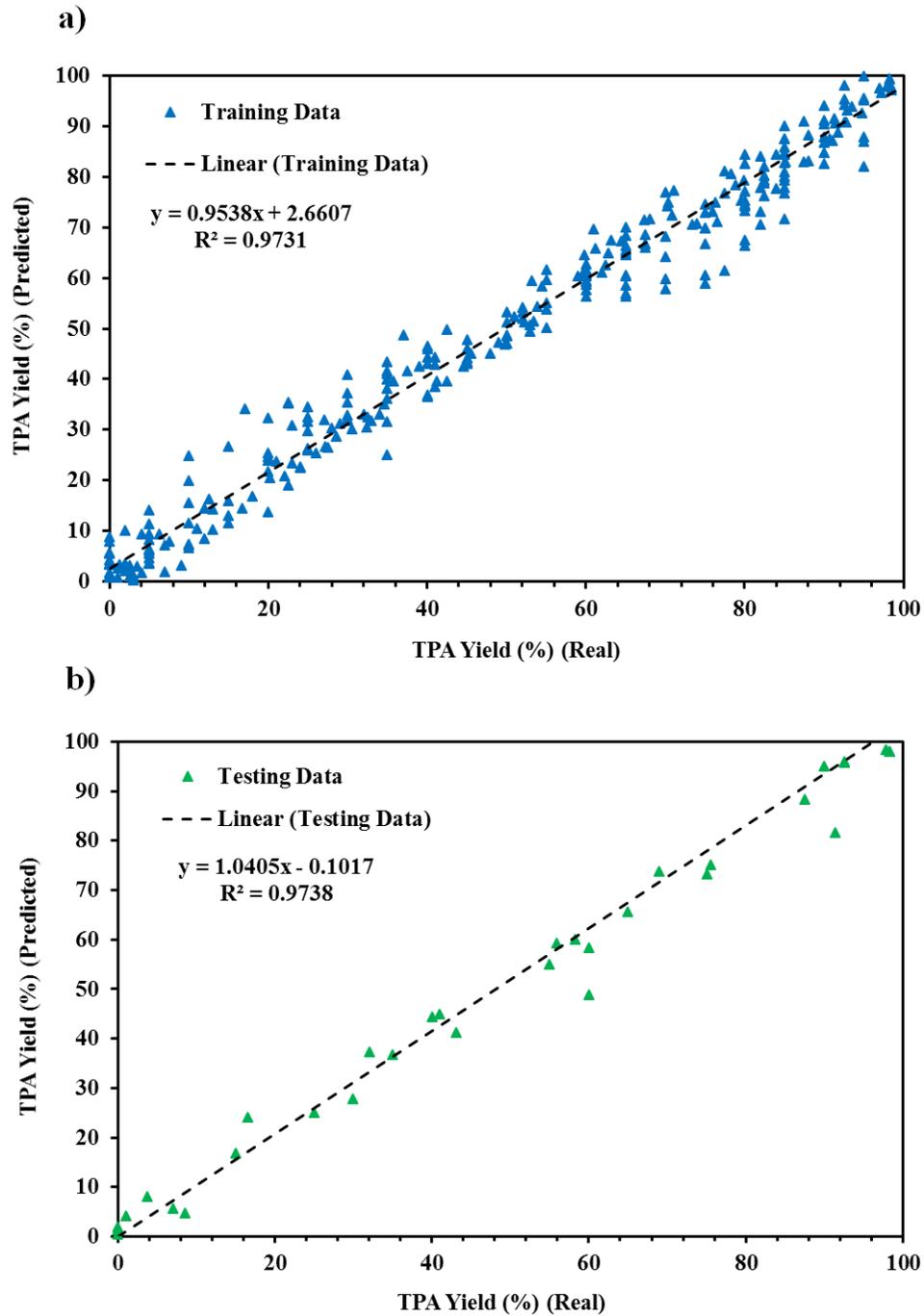

**Fig. 11 Regression plots for the TPA yield (%) prediction applying ANFIS model: a) training set, b) testing set**

**Conclusions**

In this work, a dataset was prepared based on the reaction conditions, input variables, and the produced product, TPA yield, for aqueous hydrolysis of PET. The p-value calculations confirmed considered input variables are effective on hydrolysis. The logistic regression was applied to rank the input variables where the reaction temperature and catalyst type were determined to be the most and least effective input variables for hydrolysis. The MLP-ANN and the ANFIS models were applied to anticipate TPA yield for the dataset. For that, the dataset was divided into training set and testing set. Then, they were fed into the two models to train and test their performance for comparison. For the testing data, $R^2$ and RMSE were 0.9542 and 7.81 for the MLP model, respectively while $R^2$ and RMSE were 0.9738 and 6.54 for the ANFIS model, respectively. The results also indicated that the ANFIS model could improve the performance by 2%. So, the ANFIS model could predict TPA yield better when it encounters the unseen data. This work can open a new chapter on the machine learning applications in polymer recycling science. The results can help material scientists to have an in-hand model to predict the outcome for the novel catalysts they design for the aqueous hydrolysis of PET in a safe environment.